\documentclass{article}
\usepackage{ijcai24}

\usepackage{times}
\usepackage{soul}
\usepackage{url}
\usepackage[hidelinks]{hyperref}
\usepackage[utf8]{inputenc}
\usepackage[small]{caption}
\usepackage{graphicx}
\usepackage{amsmath,amsfonts,amssymb}
\usepackage{amsthm}
\usepackage{booktabs}
\usepackage{algorithm}
\usepackage{algorithmic}
\usepackage[switch]{lineno}
\usepackage{multirow}
\usepackage{makecell}
\title{KALE: An Artwork Image Captioning System Augmented with Heterogeneous Graph}

\author{
Yanbei Jiang
\and
Krista A. Ehinger\and
Jey Han Lau\\
\affiliations
School of Computing and Information Systems, The University of Melbourne\\
\emails
yanbeij@student.unimelb.edu.au,
kris.ehinger@unimelb.edu.au,
laujh@unimelb.edu.au
}
\begin{document}

\maketitle

\begin{abstract}
    Exploring the narratives conveyed by fine-art paintings is a challenge in image captioning, where the goal is to generate descriptions that not only precisely represent the visual content but also offer a in-depth interpretation of the artwork's meaning. The task is particularly complex for artwork images due to their diverse interpretations and varied aesthetic principles across different artistic schools and styles. In response to this, we present KALE (\textbf{K}nowledge-\textbf{A}ugmented vision-\textbf{L}anguage model for artwork \textbf{E}laborations), a novel approach that enhances existing vision-language models by integrating artwork metadata as additional knowledge. KALE incorporates the metadata in two ways: firstly as direct textual input, and secondly through a multimodal heterogeneous knowledge graph. To optimize the learning of graph representations, we introduce a new \textit{cross-modal alignment loss} that maximizes the similarity between the image and its corresponding metadata. Experimental results demonstrate that KALE achieves strong performance (when evaluated with CIDEr, in particular) over existing state-of-the-art work across several artwork datasets. Source code of the project is available at \url{https://github.com/Yanbei-Jiang/Artwork-Interpretation}.
    %Our system overcomes the limitations of existing models that are designed only for natural images and contributes to the development of a new direction for image captioning research that considers the unique characteristics of artwork images. However, our model struggles with some issues like incorrect attributions and over-reliance on the metadata, and its performance enhancements from knowledge graph embeddings are not universally consistent across all contexts, highlighting the intricate nature of artwork image captioning. 
\end{abstract}

\section{Introduction}
Recently, research in the field of Artificial Intelligence (AI) has shown a growing interest in exploring the intersection between AI and Art. In the last few years, numerous initiatives have aimed to leverage AI technologies to make the domain of art more accessible and interpretable \cite{ma2017part,gonthier2018weakly,wynen2018unsupervised}. One such application is the generation of descriptions for visual arts, which is a case of image captioning.
%The interpretation of art is one of the most elusive domains of interest, because the skill to recognize meaning patterns in artworks is inextricably linked to human perception and bring art closer to people.
%The objective is to develop models that can bridge the gap between these two modalities effectively. Such models would enable machines to perform tasks that require both visual and linguistic understanding, 
%Art description generation can be seen as a problem called image captioning,
This task aims to automatically produce a short meaningful text given an input image. Beyond simple object and scene recognition, effective image captioning requires machines to understand the context and relationships among the elements within the image. 
%A recent evolution in this space is the emergence of pre-trained vision-language models, which have been trained on various datasets comprising both images and text, are now achieving state-of-the-art performance (cite here). In the context of artworks,
Through appropriate analysis and extraction of high-level features from artwork images, generated descriptions could potentially convey implicit meanings that artists want to express and make artworks more accessible. 

% A recent evolution in this space is utilizing pre-trained vision-language models. These models are pre-trained on large-scale, multimodal datasets containing both visual and textual elements. Training often begins with self-supervised learning on these datasets, where the models learn to identify and correlate features within the data without explicit labels. Then, the training continues with fine-tuning through supervised learning for specific downstream tasks like image captioning, visual question answering, and image-text retrieval. 
% One of the key drivers behind the rapid progress in pre-trained vision-language models is the availability of large-scale datasets such as COCO \cite{lin2014microsoft}, Visual Genome \cite{krishna2017visual}, and Conceptual Captions \cite{sharma2018conceptual}, which provide a rich source of visual and textual pairs that can be used to train models with billions of parameters.

%Despite extensive research on image captioning in recent years, most existing systems are designed for natural images that contain identifiable objects such as people, animals, and plants, and few contributions have been made to extend these approaches to more complex artwork images. 
Generating captions for artwork images is a challenging task for several reasons. Firstly, unlike natural images, artwork images may lack clear entities, such as in the case of abstract art, making it difficult for models to extract useful information from just the images. Another significant challenge is dealing with the ambiguity and subjectivity inherent in the image. Artwork images often have multiple levels of interpretation, and captions may vary significantly depending on the observer's cultural background and artistic taste. A descriptive caption generator might say ``some people under a tree in a park'', but a better captioning system designed for artworks might say ``a serene gathering in the shade that illustrates 19th-century pastoral life''.

\begin{figure}
\centering 
\includegraphics[width=\linewidth]{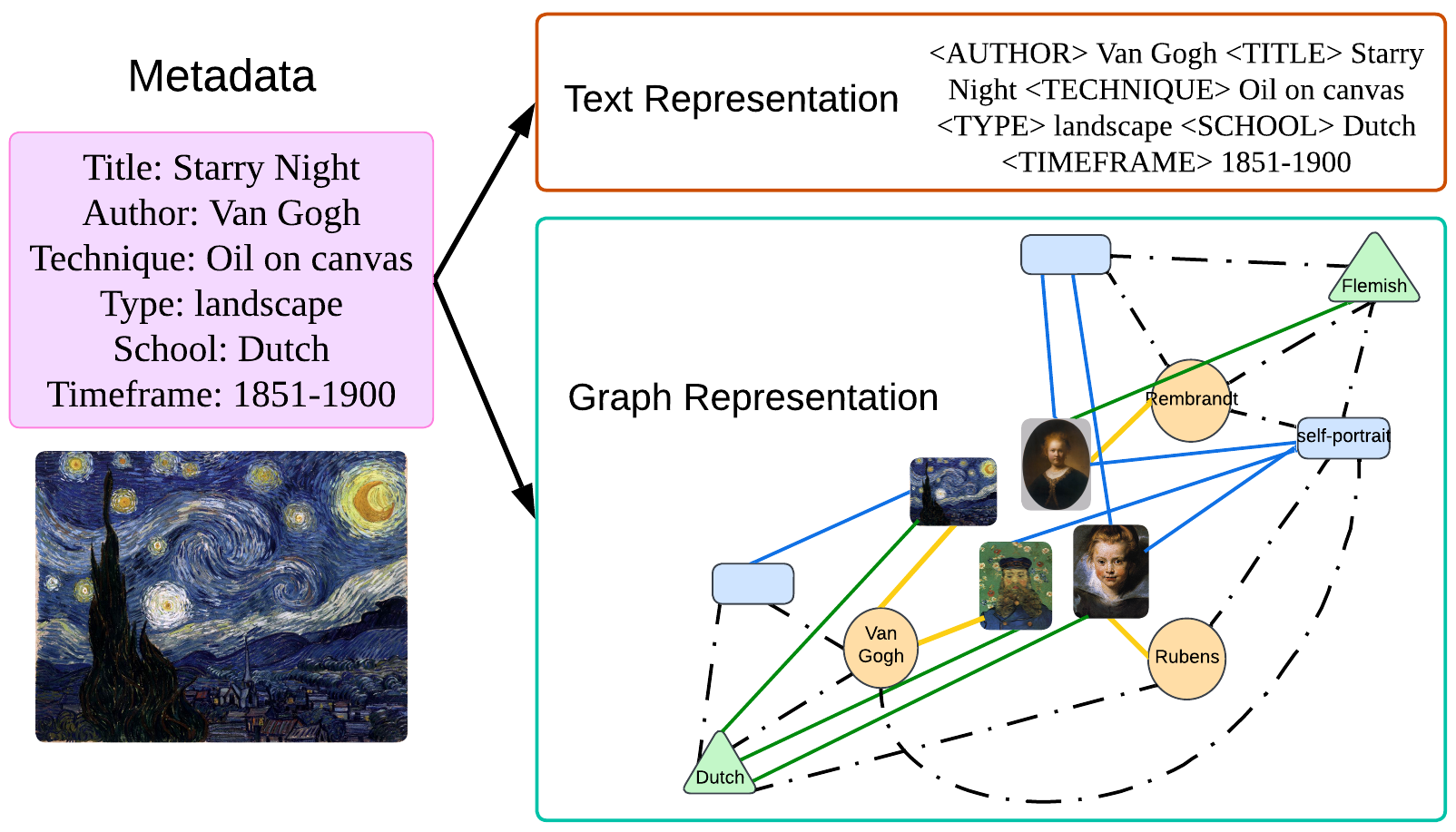}
\caption{\label{fig:intro}Artwork metadata are integrated through two ways, 1: Textual input 2: Knowledge graph input.}
\end{figure}

Recent studies, such as those by \cite{lu2022artcap}, \cite{achlioptas2021artemis} and \cite{Zixun}, employed Transformer-based encoder-decoder architectures. However, these models often lack extensive pre-training, limiting their diversity and effectiveness due to training on relatively small datasets. \cite{cetinic2021iconographic} utilizes a pre-trained model CLIP, which marks a significant advancement, but it struggles with captions that require an understanding of broader knowledge of art history. To address this, recent studies began to integrate external knowledge during training. \cite{bai2021explain} proposed a framework that utilizes external knowledge from Wikipedia, where the model detects objects in artwork images and retrieves relevant information. However, this approach has its limitations, as artwork images do not always present salient objects for detection. \cite{sheng2019generating} take a different approach by incorporating artwork types into a CNN-LSTM model, but their model is constrained by its reliance on a single external data source.

%In (cite here)'s work, artwork descriptions are broadly divided into two categories, namely visual and contextual descriptions. Visual descriptions focus on the visual elements of the artwork, such as the objects and scenes depicted. Contextual descriptions, on the other hand, often incorporate high level of interpretations that go beyond the image content, such as the artist’s latent expression, historical events and cultural significance, which are often challenging for models to figure out without a broader understanding of art history. (cite here)'s work introduced another way of annotating sentences, where "Form” deals with the visual composition, “Content” addresses the underlying meaning or subject matter, and “Context” provides the necessary background for informed interpretation of the artwork.

% Consequently, while existing image captioning models trained on natural images may work well for generating visual descriptions for artwork images, they may be less effective for generating contextual descriptions. 

To handle the above limitations, our solution is to incorporate supplementary data that provides broader knowledge beyond the image. The SemArt artwork dataset \cite{garcia2018read} offers a valuable resource as it enriches each image with additional metadata. As shown in Figure \ref{fig:intro}, each image is associated with six metadata, which provide useful background information about the artworks. For example, the “School” metadata could be used to infer the artist’s style, while “Type” tells us about the format of the artwork. To incorporate these supplementary data into an image captioning system, our first approach concatenates these metadata into a word sequence and feed them as additional input. Our second approach constructs a heterogeneous knowledge graph that integrates artwork metadata with the images to build continuous representations for the metadata. These continuous representations can then be injected as additional input. 
%We utilize a Heterogeneous Attention Network \cite{wang2019heterogeneous} to effectively aggregate information from neighbouring nodes within this graph. This approach enables us to extract relevant and contextually rich embeddings that encapsulate both the visual and informational attributes of the artwork. Although the notion of heterogeneous knowledge graph has been proposed for several years, to the best of our knowledge, none of the previous artwork image captioning systems have tried to incorporate them. 
In summary, our main contributions are listed as follows:

%In a previous study, metadata were utilized to conduct cross-modal image-text retrieval tasks through the construction of a knowledge graph from metadata \cite{garcia2020contextnet}. A knowledge graph is a way of storing and organizing data that uses a graph-structured model to represent real-world entities and their relationships. This study demonstrates the efficacy of these metadata in image-text retrieval tasks. Another study enhanced an existing image captioning model by integrating one of the five metadata, namely artwork Type, into the model and obtained encouraging results \cite{sheng2019generating}. Thus, integrating this additional information could potentially aid the model in learning more implicit contextual information. 
% \subsection{Contributions}
% \textbf{An artwork story teller: a novel artwork-specific image captioning system that adapts pre-trained vision-language model}

\begin{itemize}
\item We propose KALE, an artwork image captioning system that extends the existing pre-trained vision-language model to the art domain. Through empirical evaluations, we have demonstrated that KALE largely outperforms existing artwork image captioning models across several artwork datasets.

% \textbf{Incorporate artwork metadata into our system through textual input and knowledge graph embeddings}

\item We design two ways to incorporate artwork metadata into our system. Firstly, by including all of them as additional textual inputs, and secondly, through the construction of a heterogeneous knowledge graph by treating each image and its associated metadata as distinct node types. This methodology leverages the strengths of heterogeneous graph structures and bridges the gap between visual and textual data in artwork analysis. The results suggest that the model can foster a better understanding of the narratives behind fine art pieces.

\item KALE is trained with two objectives that maximize the likelihood of generating the ground-truth caption and the similarity between the image and its corresponding metadata in the knowledge graph. 
\end{itemize}

\section{Related Work}
\paragraph{Pre-trained Vision-Language Model.}Motivated by the success of pre-trained language models like BERT \cite{devlin2019bert} in natural language processing, pre-trained vision-language models have attracted significant attention in the multi-modal domain and they have shown remarkable performance for image captioning task \cite{radford2021learning,wang2021simvlm,wang2022ofa,li2022mplug,li2020oscar}. 
%One of the key drivers behind the rapid progress in pre-trained vision-language models is the availability of 
Pre-training on large-scale datasets such as COCO \cite{lin2014microsoft}, Visual Genome \cite{krishna2017visual}, and Conceptual Captions \cite{sharma2018conceptual} allows transferring knowledge to downstream tasks with limited data and enables the models to recognize and understand a broader range of objects and contexts. Typically, these models are composed of four main components: a vision backbone used to extract features from an input image, a language backbone to process an input text, a fusion encoder that captures the intricate interactions between visual and linguistic elements, and a language decoder to generate captions.

\paragraph{Knowledge Graph.}
In recent years, the concept of knowledge graphs has gained increasing attention in the field of artificial intelligence, as they can be used to represent a wide range of information, from simple facts and relationships to complex entities and events. %Fundamentally, it can be understood as a graph-based database denoted by $G = (V, E)$, where nodes $V$ correspond to entities and edges $E$ signify relationships between these entities. 
Recent studies on incorporating knowledge graphs into image captioning systems demonstrate a significant enhancement. One approach involves using scene graphs to represent structural relationships in images \cite{yang2019auto}. Other methods, like the one proposed by \cite{zhao2023boosting}, construct multi-modal knowledge graphs that associate visual objects with named entities to generate more informative and accurate captions. In the field of artwork analysis, \cite{garcia2020contextnet} created an art-specific graph that connects paintings with their related attributes and incorporated it into cross-model retrieval task.

\paragraph{Heterogeneous Knowledge Graph.} Unlike traditional graphs which focus on homogeneous nodes and edges, Heterogeneous Knowledge Graph (HKG) includes a variety of node and edge types, allowing for the representation of multifaceted data from different domains. For instance, in the context of artworks, nodes could represent images, artists, artwork types, schools, or historical periods. Edges, meanwhile, could denote relationships such as ``created by'', ``belongs to'' or ``influenced by''. This rich structuring could represent the art world and capture some complex historical, cultural, and stylistic factors. Heterogeneous Attention Networks (HANs) are a recent innovation leveraging the richness of HKGs \cite{wang2019heterogeneous}. HANs apply the attention mechanism selectively across different types of nodes and relationships in an HKG. The output for HAN is a set of graph embeddings. These embeddings are crucial as they translate the entities and relations present in the graph into a low-dimensional, dense, and continuous vector space, which could be trained in an end-to-end manner. One of the key concepts in HKGs is the ``meta-path'', which is a sequence of relations defining a composite relationship among multiple types of entities. They enable the extraction of complex, higher-order relationships by traversing different types of nodes and edges in a sequence and capture a specific kind of interaction or relationship within the graph.

\section{Proposed Methods}
% \subsection{Preliminaries}
% The task of artwork image captioning involves formulating a methodology to generate descriptive text for a given artwork image. This process can be conceptualized as a sequence generation problem where the objective is to produce a sequence of words \( W = \{w_1, w_2, \ldots, w_N\} \) that describes the content and context of the artwork. Each word in the sequence is selected to maximize its conditional probability given the image and the preceding words in the caption.

% Mathematically, the probability of generating a caption can be expressed as the product of conditional probabilities of each word in the sequence:

% \[ P(W|I) = \prod_{i=1}^{N} P(w_i|I, w_1, w_2, \ldots, w_{i-1}) \]

% Here, \( I \) represents the input artwork image, and \( P(w_i|I, w_1, w_2, \ldots, w_{i-1}) \) denotes the probability of the \( i^{th} \) word given the image and all preceding words \( \{w_1, w_2, \ldots, w_{i-1}\} \) in the caption. The goal is to construct a caption where this probability is maximized, effectively capturing the most relevant and coherent description for the image.

\subsection{Heterogeneous Graph Construction}
\label{section:Heterogeneous Graph Construction}
The construction of our heterogeneous graph commences with the definition of nodes and edges. In this graph, the diversity of node types is a key feature. We aim to create a multidimensional representation of the artwork and have a deeper analysis and interpretation through the graph. The nodes include:
\begin{itemize}
\item \textbf{Artwork Image:} Includes all the images in the training set, and represents the visual component of the artwork and providing a link to each metadata node.
\item \textbf{Author:} Represents the creators of the artwork, such as Vermeer and Van Gogh. Authors are key to understanding the stylistic and historical context of a piece.
\item \textbf{Title N-grams:} To capture the essence of each artwork title, we include a range of N-grams as nodes – specifically 1-gram, 2-gram, and 3-gram, and select the most common ones, which represent the crucial keywords or phrases that characterize each artwork.
\item \textbf{Title Cluster:}  We further enrich the textual dimension by incorporating Title Clusters. We first use SentenceBert \cite{reimers2019sentence} to process the titles and use k-means with cosine similarity to create these clusters. 
\item \textbf{Technique:} Describes the methods and materials used in creating the artwork, such as \textit{oil on canvas}, which are crucial for understanding the artwork's texture and style. This metadata also include specific details about an artwork's dimensions, such as $167\times124cm$. However, they often do not contribute meaningful insights into the artwork's style, so we use regular expressions to filter out these dimension data.
\item \textbf{Type:} Represents the genre or category of the artwork, such as \textit{portrait} and \textit{still-life}. 
\item \textbf{School:} Denotes the group the artwork is associated with, such as French and Spanish, offering cultural and historical relevance.
\item \textbf{Timeframe:} Describes the era or period in which the artwork was created, such as 1650-1700, aiding in historical contextualization.

\end{itemize}
We decide to exclude the metadata \textbf{Date} as it typically provides very specific year information and so has limited utility in the context of caption generation.

\begin{figure}
\centering 
\includegraphics[width=0.95\linewidth]{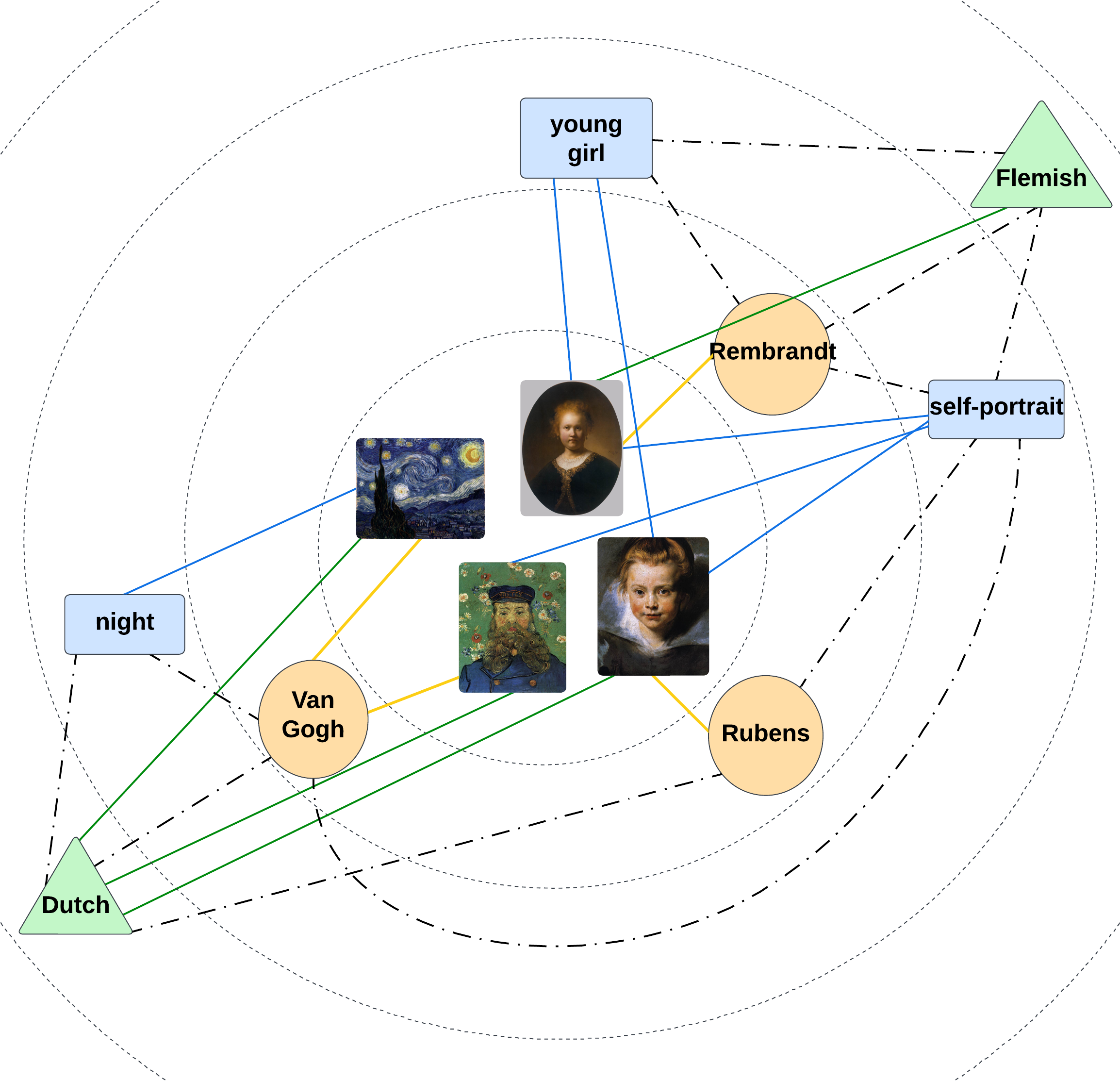}
\caption{\label{fig:graph}An example sub-graph of our multi-layer heterogeneous graph; only artwork, author (yellow), n-gram (blue) and school (green) nodes are included. For each type of nodes, it forms a layer. Solid colored lines denote image-metadata relationships, while dot-dashed lines represent inter-metadata linkages.}
\end{figure}

As depicted in Figure \ref{fig:graph}, our approach results in a multi-layered graph structure, where the artwork images act as the central nodes and form the innermost layer of the graph. Branching outward from this core are the various types of metadata, each constituting its own layer, which could be directly linked to the central artwork nodes. Edges in this graph are designed to connect across different layers. Each artwork image node is connected to all its associated metadata (shown as coloured line) and edges are also established between different types of metadata if they belong to the same artwork (shown as dot-dash line). Note that there are no direct edges within layers but connections could be established through meta-paths. The meta-path enables indirect connections between one type of nodes, providing a means to uncover deeper insights and relationships in the graph. Here are some example meta-paths we defined in the graph:

\begin{itemize}
\item \textbf{Artwork-Author-Artwork:} This meta-path connects artworks through their authors. It can be used to explore the range and diversity within an individual author's body of work.
\item \textbf{Type-Ngrams-Type:} This meta-path links art types through common themes found in artwork titles, suggesting shared or overlapping concepts between two types.
\item \textbf{Artwork-Timeframe-Artwork:} This path connects different artworks based on the time period in which they were created, allowing for analysis of historical trends or evolution in art.
\end{itemize}
The last step is to create initial embeddings for nodes. We use ResNet50 \cite{he2016deep} to extract embeddings for images. For textual data such as Author, Title N-grams, and Technique, we use pre-trained FastText model \cite{joulin2016bag}. For the Title Cluster nodes, we derive the embeddings by using the centroids of each cluster. For categorical data like Type, School, and Timeframe, we use one-hot encoding. The choice of pre-trained node embeddings is flexible in our architecture, and in future work it would be interesting to explore other pre-trained embeddings. In total, the resulting graph presents 28,796 nodes and 405,384 edges, with 20,310 images, 3,227 authors, 4,500 n-gram keywords, 100 clusters, 601 techniques, 26 schools, 22 timeframes and 10 types.

\subsection{The KALE Model}
\begin{figure*}
\centering 
\includegraphics[width=0.83\linewidth]{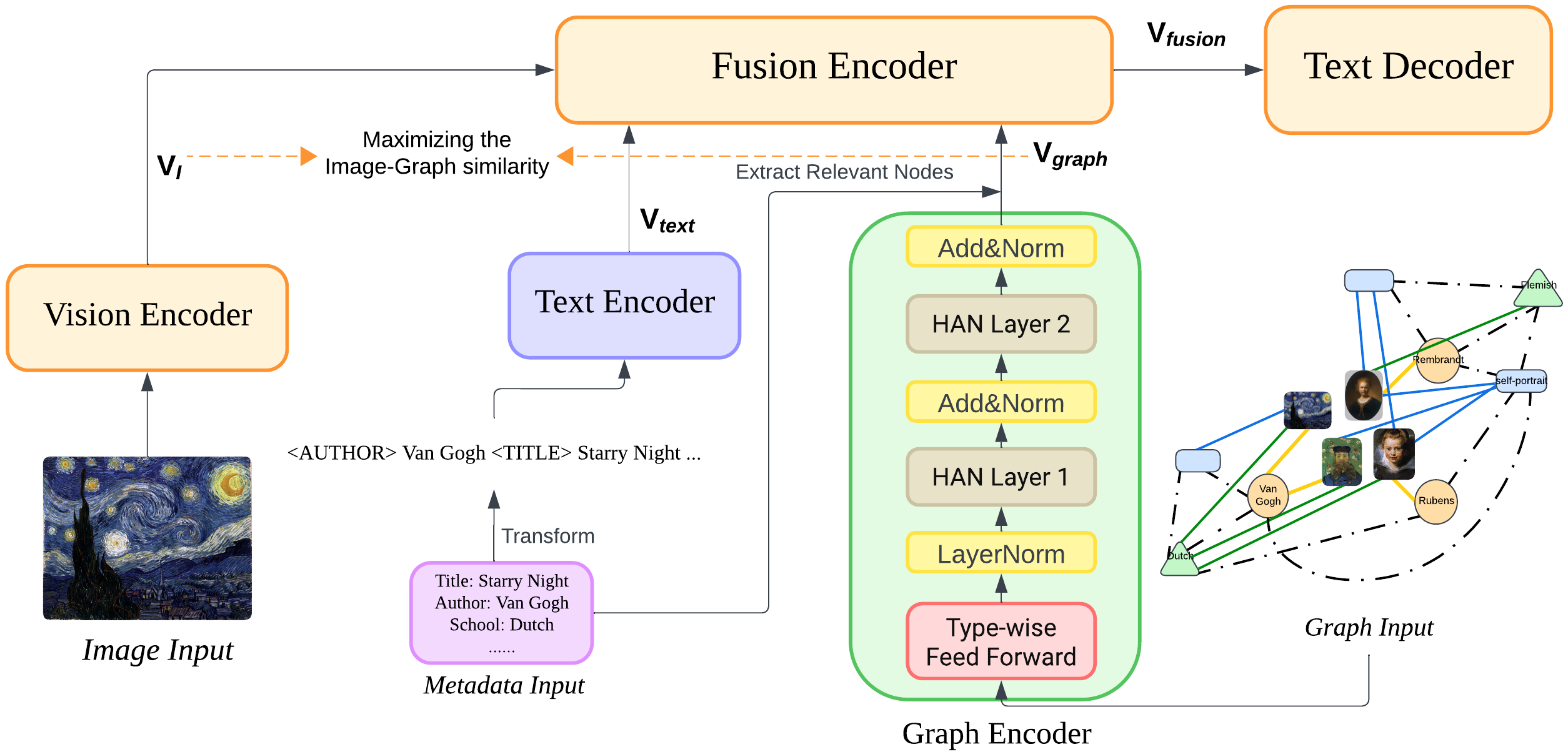}
\caption{The architecture of our KALE model. There are five main components: (1) vision encoder; (2) text encoder; (3) graph encoder; (4) fusion encoder; and (5) text decoder.
}
\label{fig:structure}
\end{figure*}

At its core, KALE extends the existing pre-trained vision-language model and incorporates artwork metadata into the model through two ways: 1) as textual input, 2) through the inclusion of knowledge graph embeddings. Figure \ref{fig:structure} depicts the general architecture for our KALE model, which has five main components: 1) Vision Encoder, 2) Text Encoder, 3) Graph Encoder, 4) Fusion Encoder, 5) Text Decoder. The general pipeline functions as follows:

Our input includes an image, artwork metadata, and a constructed heterogeneous graph with initial node embeddings. The process begins with the image passing through the vision encoder, which yields the image embedding \(\mathbf{v}_I \). Concurrently, the text encoder processes the artwork metadata to obtain the text embedding \( \mathbf{v}_{\text{text}} \). Additionally, the graph is fed into the graph encoder, from which we extract the graph embedding for related nodes, denoted as \( \mathbf{v}_{\text{graph}} \). These embeddings are then concatenated and input into the fusion encoder, which integrates the information from these three modalities into a fused embedding. Finally, the text decoder takes this fused embedding as input to generate captions. In the next few paragraphs, we introduce each of these components in detail.

\paragraph{Text Encoder.}The text encoder is used to process and encode the metadata associated with each artwork as a textual input. Our approach involves concatenating various metadata elements into a single string, with each element separated by specially designed tokens. For instance, an author's name is preceded by an $<$AUTHOR$>$ token, the title of an artwork by a $<$TITLE$>$ token, and so on. Once concatenated, this string is then fed into a pre-trained language model, BERT \cite{devlin2019bert}. This step results in the generation of text embeddings, which are vector representations capturing the semantic meaning of the metadata. Finally, we extract embeddings for only each of the special tokens and merge them to get the final textual representation. Formally, 
\begin{equation}
\mathbf{v}_{\text{text}} = [\mathbf{e}_{<\text{AUTHOR}>}; \mathbf{e}_{<\text{TITLE}>}; \mathbf{e}_{<\text{TECHNIQUE}>}; \ldots]
\end{equation}
where $\mathbf{e}_{<\text{...}>}$ indicates the embedding obtained from BERT, and $[;]$ indicates concatenation. 
%Note that as discussed in section \ref{section:Heterogeneous Graph Construction}, we also exclude the Date and second half of the Technique metadata here as they still contribute minimally as textual inputs.

\paragraph{Graph Encoder.}
The graph encoder is responsible for learning the embeddings for the nodes in the heterogeneous graph. The graph includes nodes representing different types of metadata, each with varying embedding sizes. To achieve uniformity in dimensionality, we first apply a linear layer followed by layer normalization to each node type in the graph. This step projects all node embeddings into a common dimension, referred to ``Type-wise Feed Forward'' in Figure \ref{fig:structure}. Mathematically, for a node of type $t$ with initial embedding $\mathbf{v}_\text{init}^{(t)}$ from the graph, the transformation via a linear layer can be represented as:
\begin{equation}
\mathbf{v}_\text{proj}^{(t)} = \text{FFN}_t(\mathbf{v}_\text{init}^{(t)})
\end{equation}
Next, we use two Heterogeneous Attention Network (HAN) layers to process the graph. Initially, HAN focuses on node-level attention. This process involves computing attention coefficients for each node, taking into account its neighbors, highlighting the most significant connections based on node and edge types. Mathematically,
\begin{equation}
\alpha_{ij}^P = \text{softmax}\left(\sigma\left( \mathbf{a}_P^\intercal \cdot [\mathbf{W}_P \mathbf{v}_{\text{proj}_i} \| \mathbf{W}_P \mathbf{v}_{\text{proj}_j}] \right)\right)
\end{equation}
where \( \mathbf{W}_P \) is the weight matrix under meta-path $P$, \( \mathbf{a} \) is the attention mechanism's learnable weight vector, and \( \| \) denotes concatenation. The attention coefficients \( \alpha_{ij}^P \) determine the importance of node \( j \)'s features to node \( i \). The node-level embeddings under a meta-path $P$, \( \mathbf{v}_{{i}}^P \), are computed by aggregating these weighted features:
\begin{equation}
\mathbf{v}_{i}^P = \|^K_{k=1} \sigma\left( \sum_{j \in \mathcal{N}(i)}\alpha_{ij}^P (\mathbf{W}_P \mathbf{v}_{\text{proj}_j}) \right)
\end{equation}
where \( \mathcal{N}(i) \) denotes the neighborhood of node \( i \), \( \sigma \) represents an activation function and $K$ is the number of heads in multi-head attention. HAN then extends this mechanism to a meta-path level, where it aggregates the node-level embeddings across different meta-paths or ``relations'' as different relations have their own parameters. Meta-path attention is similar to the node-level attention, the final graph embeddings, \( \mathbf{v}_{\text{all}} \), are computed as:
\begin{equation}
e^p = \text{softmax}\left( \frac{1}{|S|} \sum_{i \in S} \left( \mathbf{q}^\intercal \cdot \tanh(\mathbf{W} \cdot \mathbf{v}_{i}^P + \mathbf{b})\right) \right)  
\end{equation}

\begin{equation}
\mathbf{v}_{\text{all}} = \sum_{P} e^P \cdot \mathbf{v}^P 
\end{equation}
where $\mathbf{q}$, $\mathbf{W}$ and $\mathbf{b}$ are learnable parameters shared over all paths, $S$ represents all the nodes in the graph and $e^p$ represents the importance of meta-path $P$.
% \begin{equation}
% \alpha_{P} = \text{Attention}_{\text{meta-path}}(\mathbf{v}_{\text{node}}^P, P)
% \end{equation}
% \[ \mathbf{v}_{\text{context}} = \sum_{P} \alpha_{P} \mathbf{v}_{\text{node}}^P \]

In the final stage, we extract the corresponding nodes in the graph based on the current input metadata and concatenate them to get the final graph embedding. 
\begin{equation}
\mathbf{v}_{\text{graph}} = [\mathbf{v}_{\text{node}_1}; \mathbf{v}_{\text{node}_2}; \ldots], \text{where } 
 \mathbf{v}_{\text{node}_i} \in \mathbf{v}_{\text{all}}
\end{equation}
Note that during the test phase, sometimes the metadata may not be present in the training graph. To address this, for unseen Author, Title N-grams, Technique, we use FastText followed by the ``Type-wise Feed Forward'' layer in graph encoder to transform such metadata into embeddings. For Type, School, and Timeframe, we initialize to zero vectors.

\paragraph{Vision Encoder, Fusion Encoder, Text Decoder.}
In KALE the vision encoder, fusion encoder, and text decoder components are based on the architecture and weights of the pre-trained vision-language model, mPLUG, which achieved state-of-the-art performance in standard image captioning tasks \cite{li2022mplug}. As we are adapting vision-language models to
to a new domain (i.e.\ from natural images to artwork images), we chose mPLUG given it was pretrainede on a rich set of image-text pairs which is likely to align well with our domain. Moreover, it uses a Vision Transformer (ViT) \cite{dosovitskiy2020image} which  processes the input image by dividing it into a grid of regular patches, which is a more sensible approach for artwork images which often lack clear entities. 

\paragraph{Multi-Task Training.}
We use multi-task learning to fine-tune our model. The first task utilizes the cross-entropy loss function, which is a standard approach in image captioning problems. Mathematically, this can be expressed as:
\begin{equation}
L_{CE} = -\sum_{i=1}^{N} y_i \log(\hat{y}_i)
\end{equation}
The second task introduces a cross-modal alignment loss. This loss function is designed to maximize the cosine similarity between the image embedding and all its corresponding metadata graph embeddings during training.

For the metadata graph embeddings $\mathbf{v}_{\text{graph}}$, which are obtained from the graph encoder, we pass them through a linear layer to project into the same space as the image. For the image embedding $\mathbf{v}_I$ from the vision encoder, we use a max pooling operation, which helps to reduce the dimensionality of the embeddings while preserving the most salient features. Formally, 
\begin{equation}
\mathbf{v}_M = \text{FFN}(\mathbf{v}_{\text{graph}})
\end{equation}
\begin{equation}
\mathbf{v}_{I_{MP}} = \text{MaxPooling}(\mathbf{v}_I)
\end{equation}
The loss function for this task can be represented as:
\begin{equation}
L_{CMA} = 1 - \frac{\mathbf{v}_{I_{MP}} \cdot \mathbf{v}_M}{\|\mathbf{v}_{I_{MP}}\| \|\mathbf{v}_M\|}
\end{equation}
To combine the two loss functions, we introduce a balancing parameter $\beta$. The combined loss function can be expressed as:
\begin{equation}
L_{total} = (1 - \beta) \times L_{CE} + \beta \times L_{CMA}
\end{equation}
\section{Experiments and Results}
\subsection{Experiment Setup}
\paragraph{Datasets and Competitors.}
Four artwork datasets are considered as benchmarks in this work, they are Artpedia, SemArt v1.0, SemArt v2.0 and ArtCap. Artpedia comprises a collection of 2,930 paintings from the 13th to the 21st century, each of which is associated with some textual descriptions and a corresponding title \cite{stefanini2019artpedia}. SemArt dataset was first proposed by \cite{garcia2018read} for cross-modal retrieval tasks, and contains European fine-art reproductions from the Web Gallery of Art. However, the descriptions of artworks are lengthy paragraphs, which do not align well with the image captioning task. To solve this, one recent study \cite{Zixun} separates each paragraph into single sentences, and labels them as visual and contextual sentences. The former describes the simple visual appearance of the artwork and the latter provides information about the painting’s historical background. We call this as SemArt v1.0. \cite{bai2021explain} proposed another more fine-grained way of categorising these sentences, based on their Form, Content and Context. Form deals with the visual composition, Content addresses the underlying meaning or subject matter, and Context provides the background of the artwork. We call this as SemArt v2.0 dataset. The last dataset ArtCap, contains 3605 paintings from WikiArt and the captions were manually collected by crowd-sourcing \cite{lu2022artcap}.

We consider three previous state-of-the-art models as our baselines. \cite{Zixun} applies the Meshed-Memory transformer \cite{cornia2020meshed} to the artwork domain. We refer to this work as \textit{Wu2022}. \cite{bai2021explain} proposed a framework incorporating external knowledge by leveraging a knowledge retriever from Wikipedia and training a knowledge-filling module as a ``fill in the blank'' task to incorporate art information relevant to each painting. We refer to this work as \textit{Bai2021}. For the last baseline, \cite{lu2022artcap} proposed a virtual-real semantic alignment training process through generating a virtual painting dataset via style transfer and  training a painting feature extractor using the virtual dataset with a semantic alignment loss. We refer to this work as \textit{Lu2022}.

\paragraph{Implementation details.}
For the vision encoder, fusion encoder and text decoder, we follow by default settings of mPLUG$_{large}$ model given in the open source code.\footnote{\url{https://github.com/alibaba/AliceMind/tree/main/mPLUG}}

%For instance, mPLUG$_{large}$ contains 6 text encoder layers, 6 fusion encoder layers, 12 text decoder layer and use ViT-L-14 as visual encoder. 
For the heteregeneous graph construction, the cluster number of title embedding is set to 100, 1-gram of title is set to 2000, 2-gram is 1500 and 3-gram is 1000. 
%We use two layer HAN to process the graph embedding, the attention head number is set to 8 for the first HAN layer and 4 for the second HAN layer. And to match the hidden size of fusion encoder, we set the output size of HAN to be 768. 
For the optimizer, after parameter tuning on validation set, we use AdamW \cite{loshchilov2018decouple} optimizer with a weight decay of 0.02. The learning rate is first warmed up to 5e-5 for vision encoder, 1e-2 for graph encoder, and 1e-4 for other layers in the first 1000 iterations and decayed to 1e-5 following a cosine schedule. We use beam search decoding with beam width 5 and $\beta$ is set to 0.2 to balance two loss functions.

\begin{table*}
\centering
%\renewcommand\arraystretch{1.4}
%\resizebox{0.8\textwidth}{!}{
\resizebox{0.82\textwidth}{!}{
\begin{tabular}{cccccccccc}
\toprule
\multicolumn{1}{l}{} &
\multicolumn{1}{c}{} &
\multicolumn{8}{c}{\textbf{Evaluation Metrics}} \\
\cmidrule(lr){3-10} 
\makecell{\textbf{Dataset}} & \textbf{Models} & C & B-1 & B-2 & B-3 & B-4 & M & S & R \\

\midrule      

\multirow{3}{*}{Artpedia}
                                        & Wu2022 & 3.94 & 24.7 & - & - & 3.06 & 6.58 & - & 22.4 \\
                                        & KALE (w/o metadata) & 11.7 & 29.9 & 15.0 & 7.95 & 4.77 & 8.02 & 5.49 & 22.4 \\
                                        & KALE (w/ metadata) & \textbf{23.4} & \textbf{32.6} & \textbf{17.7} & \textbf{10.9} & \textbf{7.48} & \textbf{9.33} & \textbf{7.68} & \textbf{23.7} \\
\midrule

\multirow{2}{*}{ArtCaps} 
                                        & Lu2022 & 15.21 & 52.89 & 33.11 & 20.42 & 12.57 & 15.38 & 8.39 & 36.15\\
                                        & KALE (w/o metadata) & \textbf{36.51} & \textbf{59.78} & \textbf{41.24} & \textbf{28.56} & \textbf{20.14} & \textbf{19.39} & \textbf{12.06} & \textbf{42.22} \\
\midrule

\multirow{3}{*}{\makecell{SemArt v1.0 \\ (Visual)}}
                                        & Wu2022 & 6.93 & 19.20 & - & - & 3.24 & 6.28 & - & 21.90 \\
                                        & KALE (w/o metadata) & 21.34 & \textbf{29.58} & 15.60 & 9.69 & 7.17 & 8.83 & 7.23 & 22.50 \\
                                        & KALE (w/ metadata) & \textbf{29.96} & 28.68 & \textbf{15.81} & \textbf{10.42} & \textbf{7.96} & \textbf{9.08} & \textbf{8.22} & \textbf{22.58} \\
\midrule
\multirow{3}{*}{\makecell{SemArt v1.0 \\ (Contextual)}}
                                        & Wu2022 & - & - & - & - & - & - & - & - \\
                                        & KALE (w/o metadata) & 13.01 & 30.37 & 15.38 & 9.61 & 7.27 & 8.07 & 5.26 & 20.36 \\
                                        & KALE (w/ metadata) & \textbf{21.76} & \textbf{35.35} & \textbf{20.33} & \textbf{14.15} & \textbf{11.27} & \textbf{10.16} & \textbf{7.25} & \textbf{23.41} \\
                                        \midrule
\multirow{3}{*}{SemArt v2.0}
                                        & Bai2021 & 8.80 & - & - & - & \textbf{9.10} & \textbf{11.4} & - & \textbf{23.1} \\
                                        & KALE (w/o metadata) & 13.60 & 25.85 & 13.75 & 8.78 & 6.70 & 7.48 & 5.97 & 19.86 \\
                                        & KALE (w/ metadata) & \textbf{20.7} & \textbf{27.7} & \textbf{15.7} & \textbf{10.8} & 8.57 & 9.51 & \textbf{7.31} & 21.9 \\
\bottomrule
\end{tabular}}
\caption{Evaluation results on datasets Artpedia, ArtCaps, SemArt v1.0 and SemArt v2.0 over baselines. \textbf{Bold} representes the highest score. As we directly use the reported results for the competitors, some of the metrics are missing, we use “-” to indicate missing values.} 
\label{table:caption2}
\end{table*}

\subsection{Results and Analysis}
Table \ref{table:caption2} outlines the performance results on the four datasets Artpedia, ArtCaps, SemArt v1.0 and SemArt v2.0. We compare our model against three baseline models, Wu2022, Lu2022 and Bai2021. All models use the same train/validation/test split. Baseline numbers are values from the original publications.

Note that for SemArt v1.0 dataset, the author only conducts their experiments on visual sentences, so we break it into visual and contextual separately in the table. These models are evaluated over eight evaluation metrics, including CIDEr (C) \cite{vedantam2015cider}, BLEU-1 (B-1), BLEU-2 (B-2), BLEU-3 (B-3), BLEU-4 (B-4) \cite{papineni2002bleu}, METEOR (M) \cite{banerjee2005meteor}, SPICE (S) \cite{anderson2016spice} and ROUGE-L (R) \cite{lin2004rouge}.

%There are several versions of our KALE model in the table. 
We use ``KALE (w/o metadata)'' to denote the model without any metadata input, where the text and graph encoders are removed. In this case, it's an off-the-shelf pre-trained mPLUG fine-tuned on an artwork's dataset using standard generation cross-entropy objective.

%We involve it because we want to quantitatively evaluate how metadata integration enhances the model's capability. 
On the other hand, ``KALE (w/ metadata)'' represents the model with both textual and graph representations. One exception is that for Artpedia dataset, the only metadata we have is artwork title, so we treat the artwork title as textual input and remove the graph encoder. As ArtCaps dataset lacks any metadata, we only use KALE without metadata approach for fine-tuning.
%we introduce ``KALE (trained on SemArt)'' where the text and graph encoders are removed, but we initialize the parameters with the model trained on the SemArt v1.0 (visual) and subsequently fine-tune it on the ArtCaps dataset. While ArtCaps itself does not offer external knowledge, the model pre-trained on SemArt carries forward some prior knowledge, thereby transferring beneficial insights to the ArtCaps dataset. 

When we pre-process the SemArt v1.0 and v2.0 datasets, we notice some identical or highly similar image captions present in both the training set and validation/test set. Such duplication poses a risk of model overfitting, where the model might “memorize” the captions they have seen during training. Therefore, we remove such overlapping instances from the training dataset. Note that Wu2022 and Bai2021 did not do this, and so their reported performance may be an overestimate.

As evidenced by Table \ref{table:caption2}, our methodologies exhibit superiority over existing baselines. Overall, most of the best metric scores are achieved by our KALE model. By looking at the detailed scores for each metric, we find that CIDEr increases the most compared to baselines. For instance, our model achieves 23.4 CIDEr score on Artpedia dataset, which is over 5 times that of the competitor. On SemArt v2.0, our model also outperforms over 2 times than competitor, but for metrics like BlEU-4, METEOR and ROUGE, our model performs slightly worse than the competitor. Note that CIDEr offers insights into caption diversity, and this result shows that our model could generate more diverse captions. Later in our qualitative analysis, we will further verify this with some concrete examples. Looking at the performance of our KALE model without metadata, it still surpasses the Wu2022 model for Artpedia and SemArt v1.0 across all metrics. Such improvements demonstrate the effectiveness of pre-trained vision-language models in the art domain. The most important finding perhaps is the improvement of KALE leveraging metadata, which yields superior results against KALE without metadata across all metrics. This performance shows that metadata is a critical component for understanding artworks.
%The text encoder encourages a deeper understanding of semantic relationships among different metadata and the graph links metadata with each other that include spatial elements such as correlations of artworks, artists, and schools.

\begin{figure*}[ht]
\centering 
\includegraphics[width=\linewidth]{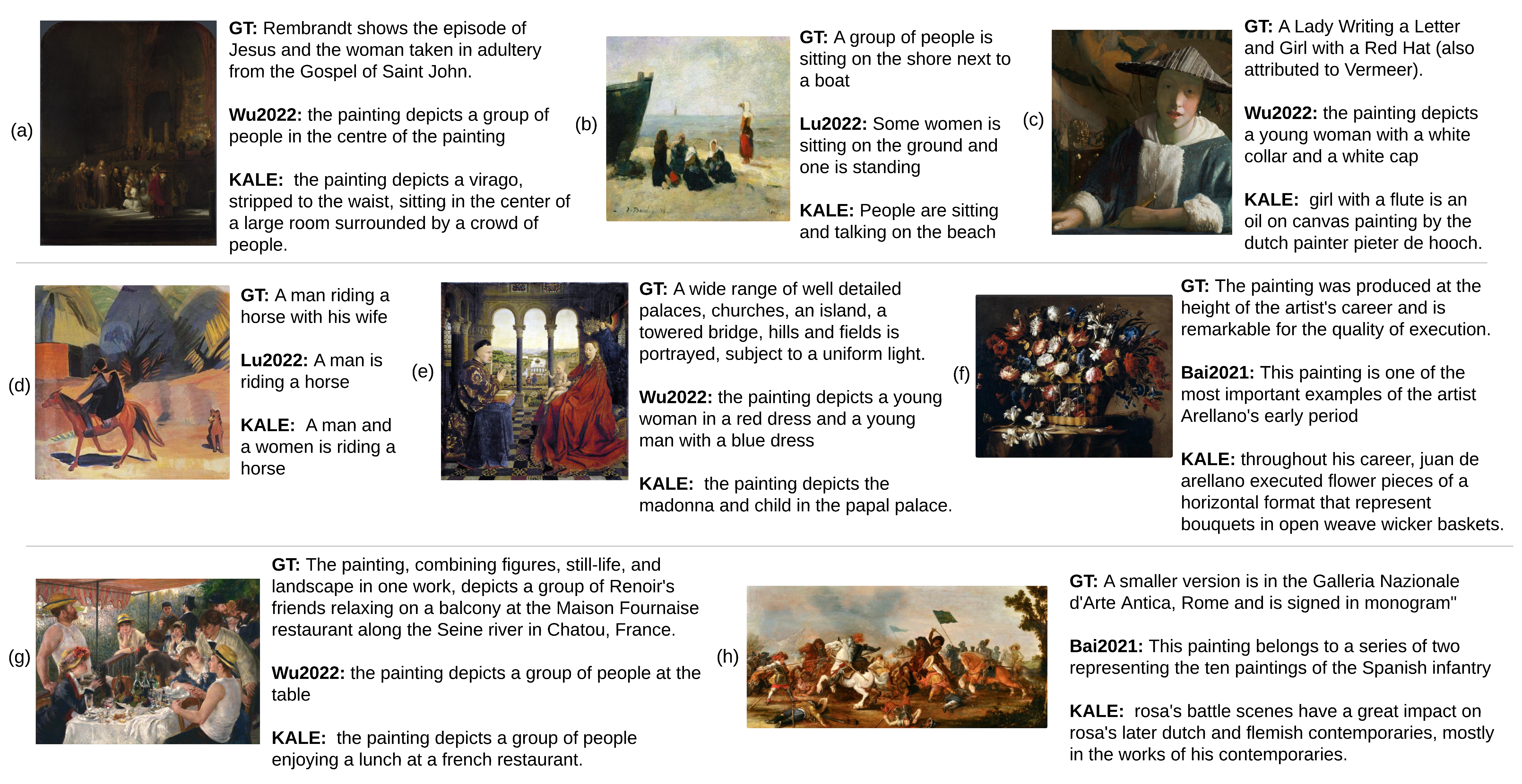}
\caption{\label{fig:example}Some examples of generated captions by KALE, along with GT (ground truth) captions and competitor-generated captions.}
\end{figure*}

\section{Discussion}
\subsection{Qualitative Analysis}
Figure \ref{fig:example} depicts several random examples from the test set of our four benchmark datasets. In terms of content correctness, KALE is able to align its captions with the visual aspects of the images in most cases. For example, in image (d), KALE recognizes there is also a woman in the image. And interestingly, sometimes it can even recognize some elements that are not immediately present in the image. For instance, in image (g), KALE identifies the setting as a ``French restaurant'' (which is correct, based on the ground truth caption (GT). %We only realize it was located at the Seine River in France after checking the ground truth caption. 
In terms of creativity, KALE's captions appear to be richer and more detailed than those of competitors. As illustrated in example (a), KALE portrays the scene as a ``virago and stripped to the waist, surrounded by a crowd of people.'' In contrast, Wu2022 only describes it as a ``group of people'', which lacks substance and depth. However, the captions produced by KALE occasionally present inaccurate details about the artwork. For instance in example (c), KALE suggests that Pieter De Hooch is the artist of the painting rather than Vermeer. Overall, the integration of metadata appears to help generate higher quality captions, especially for images that have more background context. That said, compared to the ground truth captions there is arguably still quite a bit of gap, and so there is plenty of room for improvement and artwork interpretation is by no means a solved task.

\begin{table}
\centering
\small
\resizebox{0.48\textwidth}{!}{
\begin{tabular}{cccccc}
\toprule
\multicolumn{1}{c}{} &
\multicolumn{1}{c}{} &
\multicolumn{4}{c}{\textbf{Evaluation Metrics}} \\
\cmidrule(lr){3-6} 
\makecell{\textbf{Dataset}} & \textbf{Model Variant} & C & B-1 & B-4 & M \\
\midrule
\multirow{3}{*}{\makecell{SemArt v1.0 \\ (Visual)}} 
                                            & KALE (w/o) & 21.34 & \textbf{29.58} & 7.17 & 8.83 \\
                                            & KALE (T) & 27.46 & 29.27 & 7.74 & 8.98 \\
                                            & KALE (T \& G) & \textbf{29.96} & 28.68 & \textbf{7.96} & \textbf{9.08} \\                           
\midrule
\multirow{3}{*}{\makecell{SemArt v1.0 \\ (Contextual)}} 
                                            & KALE (w/o) & 13.01 & 30.37 & 7.27 & 8.07 \\
                                            & KALE (T) & 20.35 & 34.50 & 10.78 & 9.86 \\
                                            & KALE (T \& G) & \textbf{21.76} & \textbf{35.35} & \textbf{11.27} & \textbf{10.16} \\
\midrule
\multirow{3}{*}{SemArt v2.0} 
                                            & KALE (w/o) & 13.60 & 25.85 & 6.70 & 7.48 \\
                                            & KALE (T) & 12.35 & \textbf{28.01} & 6.03 & 7.89 \\
                                            & KALE (T \& G) & \textbf{20.7} & 27.7 & \textbf{8.57} & \textbf{9.51} \\
\bottomrule
\end{tabular}}
\caption{Performance impact of text and graph on KALE across SemArt v1.0 and v2.0 datasets. KALE (w/o) refers to the version without any metadata. KALE (T) refers to the version with only textual data input. KALE (T \& G) refers to the version with both text and graph input.} 
\label{table:metadata influence}
\end{table} 

\subsection{Ablation Study}
\paragraph{Impact of Text and Graph.}
This experiment assesses the comparative impact of using metadata as only textual input versus a combination of textual and graph inputs. As presented in Table \ref{table:metadata influence}, the text-only approach demonstrates a substantial improvement in performance over most of metrics compared to the version without metadata. This indicates the significant impact that textual metadata alone can make the model generate accurate and diverse captions. Further, when KALE was augmented with both textual and graph inputs, there was an additional enhancement in its performance, especially on metrics like CIDEr, BLEU-4 and METEOR, indicating the effectiveness of the knowledge graph. Interestingly, the knowledge graph integration showed more effectiveness on datasets like SemArt v1.0 Contextual and SemArt v2.0. These datasets are characterized by many contextual sentences that demand a deeper understanding of art, a requirement that the external knowledge provided by the graph is particularly well-suited to address. 

\section{Conclusion}
In this work, we develop a novel artwork-specific image captioning system, KALE, that integrates external knowledge into the system through both text and heterogeneous graph. KALE is novel in that it captures the heterogeneity among images and several artwork attributes in the constructed graph. Results, both quantitative and qualitative, showed our method provides a better understanding of the narratives behind works of fine art. 

%Our system overcomes the limitations of existing models that are designed only for natural images and contributes to the development of a new direction for image captioning research that considers the attributes of fine-art paintings.
\clearpage

\bibliographystyle{named}
\bibliography{ijcai24}

\end{document}